\documentclass[10pt,twocolumn,letterpaper]{article}

\usepackage[arxiv]{iccv}      

\definecolor{iccvblue}{rgb}{0.21,0.49,0.74}
\usepackage[pagebackref,breaklinks,colorlinks,allcolors=iccvblue]{hyperref}

\def\paperID{751} 
\def\confName{ICCV}
\def\confYear{2025}

\title{Where am I? Cross-View Geo-localization with Natural Language Descriptions}

\author{Junyan Ye\textsuperscript{\rm 1,2}\thanks{Equal contribution.},
    Honglin Lin\textsuperscript{\rm 2*},
    Leyan Ou\textsuperscript{\rm 1}, \\
    Dairong Chen\textsuperscript{\rm 4,1},
    Zihao Wang\textsuperscript{\rm 1},
    Qi Zhu\textsuperscript{\rm 1},
    Conghui He\textsuperscript{\rm 2,3},
    Weijia Li\textsuperscript{\rm 1}\thanks{Corresponding authors.}\\
    \textsuperscript{\rm 1}Sun Yat-Sen University, 
    \textsuperscript{\rm 2}Shanghai AI Laboratory,
    \textsuperscript{\rm 3}Sensetime Research,
    \textsuperscript{\rm 4}Wuhan University
    \\
    }

\usepackage{amsmath}
\usepackage{amssymb}
\usepackage{mathtools}
\usepackage{amsthm}
\usepackage{algorithm}
\usepackage{algorithmic}

\usepackage{pifont}
\usepackage{threeparttable} 

\usepackage{multirow}       
\usepackage{amsmath}        
\usepackage{graphicx}       
\usepackage{caption}        
\usepackage{booktabs}       
\usepackage{adjustbox}      

\PassOptionsToPackage{table}{xcolor}

\usepackage{microtype}
\usepackage{graphicx}
\usepackage{booktabs} 
\usepackage{multirow}
\usepackage{makecell}

\usepackage{amsmath, amssymb, amscd, amsthm, amsfonts}
\usepackage{bm}
\usepackage{graphicx} 

\usepackage{amsmath,epstopdf,amssymb,color,url,multirow,multicol,verbatim,threeparttable,diagbox,makecell,booktabs}
\definecolor{light-gray}{gray}{0.82}
\definecolor{aliceblue}{rgb}{0.94,0.97,1.0}

\begin{document}

\twocolumn
[{%
\renewcommand\twocolumn[1][]{#1}%
\maketitle
\begin{center}
\vspace{-0.5cm}
\captionsetup{type=figure}
\includegraphics[width=0.85\linewidth]{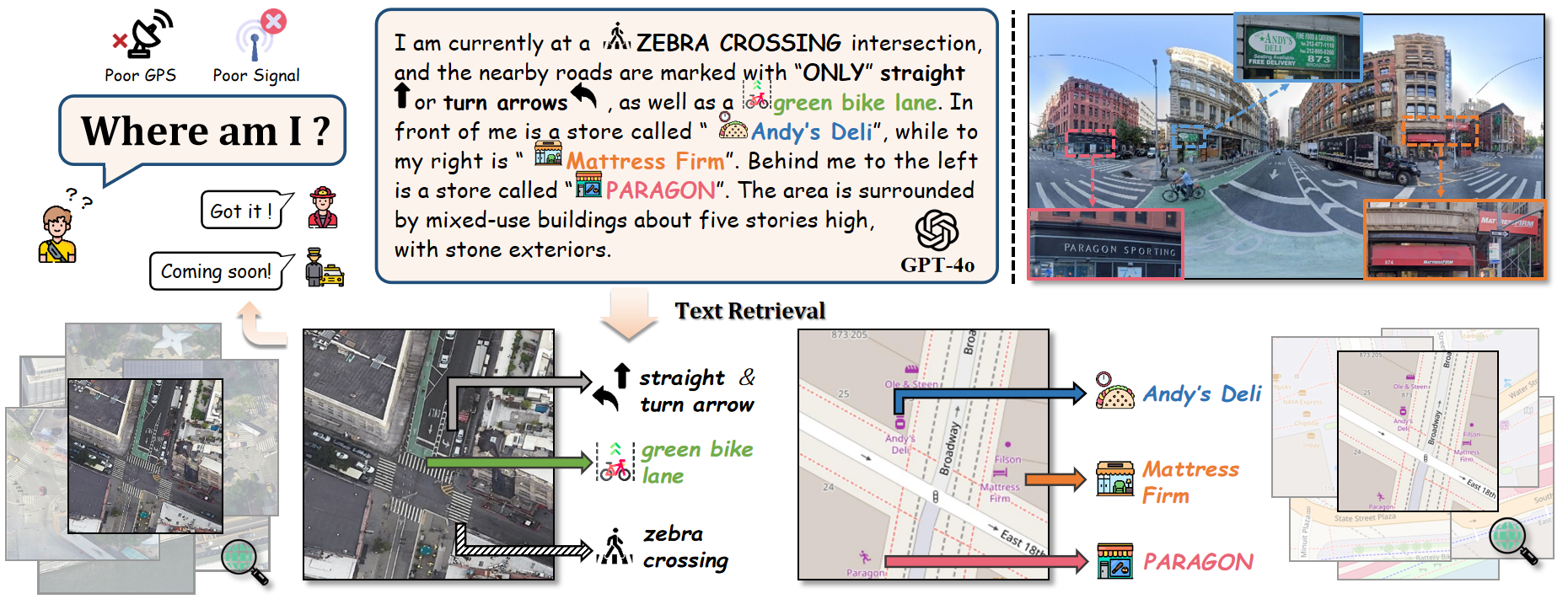}
\captionof{figure}{\textbf{Towards Text-guided Geo-localization.} In scenarios where GPS signals are interfered with, users must describe their surroundings using natural language, providing various location cues to determine their position \textit{(Up)}. To address this, we introduce a text-based cross-view geo-localization task, which retrieves satellite imagery or OSM data only based on text queries for position localization \textit{(Down)}.}
\label{fig:teaser}
\end{center}%
}]

\vspace{-20mm}
\begin{abstract}
Cross-view geo-localization identifies the locations of street-view images by matching them with geo-tagged satellite images or OSM. However, most existing studies focus on image-to-image retrieval, with fewer addressing text-guided retrieval, a task vital for applications like pedestrian navigation and emergency response.
In this work, we introduce a novel task for cross-view geo-localization with natural language descriptions, which aims to retrieve corresponding satellite images or OSM database based on scene text descriptions. 
To support this task, we construct the CVG-Text dataset by collecting cross-view data from multiple cities and employing a scene text generation approach that leverages the annotation capabilities of Large Multimodal Models to produce high-quality scene text descriptions with localization details.  
Additionally, we propose a novel text-based retrieval localization method, CrossText2Loc, which improves recall by 10\% and demonstrates excellent long-text retrieval capabilities. In terms of explainability, it not only provides similarity scores but also offers retrieval reasons. More information can be found at \url{https://yejy53.github.io/CVG-Text/}.

\end{abstract}

\section{Introduction}
\label{sec:intro}

Accurate positioning of ground-level images is crucial for various applications, including pedestrian navigation \cite{tian2024loc4plan}, mobile robot localization \cite{sarlin2019coarse}, and noisy GPS signals correction in crowded urban areas \cite{wang2024fine}. Traditional localization methods typically rely on 3D point cloud positioning \cite{irschara2009structure,lynen2020large} or cross-view retrieval using GPS-tagged satellite images \cite{Deuser_2023_ICCV,ye2024sg,ye2024cross}. However, most research focuses on matching image-to-3D data or image-to-image data. Recent studies have begun to explore a novel localization approach based on natural language text, which holds significant value for many practical applications \cite{xia2021vpc,xia2024text2loc,tian2024loc4plan}. As shown in Figure \ref{fig:teaser}, taxi drivers rely on verbal instructions from passengers to determine their location \cite{kolmet2022text2pos}, or pedestrians describe their position during emergency calls \cite{chen2024scene}.

Recent natural language localization methods, such as Text2Pose\cite{kolmet2022text2pos} and Text2Loc \cite{xia2024text2loc}, have been limited to using natural language to identify individual locations within point clouds. Constructing 3D maps using LiDAR or photogrammetry is expensive on a global scale \cite{frahm2010building,agarwal2011building,moulon2017openmvg}, and the storage costs for 3D maps are also high, often requiring costly cloud infrastructure, which hinders localization on mobile devices. Notably, the cross-view retrieval geo-localization paradigm \cite{zhu2021vigor,xia2025adapting,ye2024cross} that utilizes OSM\footnote{\url{https://www.openstreetmap.org/}} map data or satellite imagery, although oriented towards coarse-grained localization, can still meet the needs of most tasks and has clear advantages over 3D data in terms of coverage and storage costs. Therefore, this paper introduces a novel  cross-view geo-localization task, i.e., exploring the use of natural language descriptions to retrieve corresponding OSM or satellite images.

To tackle this challenging task, a dataset is needed that (i) contains foundational cross-view data with street-view, OSM, and satellite images, and (ii) includes text data capable of simulating human users in describing street-view scenes, while providing high-quality scene localization cues. With the development of large multimodal models (LMMs), annotating text using LMMs seems to be an effective solution \cite{he2023crowdsourcing,cao2024latteclip,chen2023sharegpt4v}. However, LMMs may suffer from vague descriptions or hallucination phenomena \cite{rawte2023survey,huang2023survey}. To address these issues, we propose the Cross-View Geo-localization dataset, CVG-Text. We first collected street-view data from over 30,000 locations in three cities, New York, Brisbane, and Tokyo. Then, based on the geographical coordinates, we obtained corresponding paired data of OSM and satellite images. Subsequently, we developed a progressive text description framework that leverages LMM, GPT-4o \cite{openai2024hello} as the core for generation, combining Optical Character Recognition (OCR) \cite{mori1992historical} and Open-World Segmentation \cite{qi2022open, zheng2025open} techniques to generate high-quality scene description text from street-view images while reducing vague descriptions.

Although the textual data constructed above can provide user-like street scece descriptions, it still has still have a significant domain gap compared to satellite images or OSM data. Moreover, in order to fully capture the scene's detailed information, the generated text descriptions are generally long, often exceeding the text encoding limits of image-text retrieval methods. To address this issue, we propose a novel Cross-view Text-based Localization method, \textit{CrossText2Loc}. This method includes a length-extended text encoding module, Extended Embedding, which fully leverages the long and complex text descriptions in the dataset. Through contrastive learning strategies, it effectively learns cross-domain matching information. It also features an Explainable Retrieval Module (ERM), which provides natural language explanations alongside the retrieval results. This overcomes the limitations of traditional cross-view retrieval methods that only provide similarity scores, lacking interpretability and making it difficult to make confident decisions. We evaluate the performance of mainstream text-image retrieval methods and our method on this novel task, and the experimental results demonstrate that our \textit{CrossText2Loc} has significant advantages in recall metrics and interpretability. Our main contributions are as follows:

\begin{itemize}
\item We introduce and formalize the Cross-View Geo-localization task based on natural language descriptions, utilizing scene text descriptions to retrieve corresponding OSM or satellite images for geographical localization.
\item  We propose \textit{CVG-Text}, a dataset with well-aligned street-views, satellite images, OSM, and text descriptions across three cities and over 30,000 coordinates. Additionally a progressive scene text generation framework based on LMM is presented, which reduces vague descriptions and generates high-quality scene text.
\item We introduce \textit{CrossText2Loc}, a novel text-based localization method that excels in handling long texts and interpretability. It achieves an improvement of over 10\%  in Top-1 recall compared to existing methods, while offering retrieval reason and confidence beyond similarity scores.
\end{itemize}

\section{Related Work}
\label{sec:formatting}

\subsection{Cross-view Geo-localization}
Cross-view geo-localization identifies the geographic locations of street-view images by matching them with geographic reference satellite databases or OSM databases for coarse localization \cite{ye2024cross,shi2020looking,li2023omnicity,xia2025adapting}. For example, works like Sample4G \cite{Deuser_2023_ICCV} employ a contrastive learning framework to match features of street-view images with satellite image features, achieving high-accuracy satellite data retrieval. However, current cross-view retrieval and localization tasks primarily focus on image-to-image, with limited consideration for text, which presents certain shortcomings in practical applications. Additionally, existing cross-view retrieval methods mainly provide similarity score, with little research dedicated to confident and interpretable retrieval localization. Our proposed text retrieval localization task is based on the retrieval and localization of natural language, addressing the gap in existing research regarding text-guided scene localization applications. This approach enables more transparent and interpretable retrieval localization through the use of natural language.

\subsection{Visual Language Navigation and Localization}
Visual Language Navigation (VLN) requires agents to navigate specific environments based on natural language instructions \cite{chen2019touchdown,anderson2018vision}. Previous tasks have primarily focused on decision-making based on images and natural language. Recent works have begun to shift from visual language navigation to direct visual language localization tasks. For instance, Loc4Plan\cite{tian2024loc4plan} and AnyLoc\cite{keetha2023anyloc} emphasize the necessity of visual spatial localization prior to navigation. Text2Pose \cite{kolmet2022text2pos} and Text2Loc \cite{xia2024text2loc} explore the use of natural language to identify individual positions in outdoor point cloud maps. However, point cloud retrieval tasks tend to incur high storage and computational costs when handling large-scale areas. In contrast, our task retrieves corresponding satellite images or OSM data based on text queries, allowing for broad area coverage while reducing storage and computational costs.

\subsection{Data Synthesis via LMMs}

The rapid development of multimodal large models (LMMs) has demonstrated their outstanding ability to generate high-quality natural language descriptions \cite{openai2024hello,team2023gemini,anthropic2024claude3}. Many studies have leveraged LMMs for automated data annotation \cite{he2023crowdsourcing}, such as LatteCLIP \cite{cao2024latteclip}, which synthesizes text using LMMs for unsupervised CLIP fine-tuning. However, text retrieval localization tasks impose higher demands on the annotation capabilities of LMMs, requiring them to accurately identify key localization details in street-view images, such as store signs and other critical information, while minimizing interference from hallucination phenomena \cite{rawte2023survey,huang2023survey}. 
In our fine-grained text synthesis, we incorporate street-view image, OCR \cite{mori1992historical}, and open-world segmentation \cite{qi2022open} to enhance GPT's information capture capability and reduce hallucination. Additionally, we carefully designed system prompts to guide GPT in generating fine-grained textual descriptions based on a progressive scene analysis chain of thought.

\subsection{Multi-modality Alignment}

In this work, our task involves retrieving satellite images or OSM images based on scene text descriptions synthesized by LMMs, which can be seen as a subtask of text-to-image retrieval \cite{dou2022empirical,chu2023towards,zheng2020dual,li2019visual,yang2023towards,zeng2022multi}. The CLIP \cite{radford2021learning} model introduced a contrastive learning approach between image-text pairs, establishing a new paradigm for recent text retrieval tasks. However, such models often have fixed maximum sequence length limitations, typically defaulting to 77 tokens, making it challenging to handle complex long-text scene descriptions. This can lead to the loss of fine-grained textual information, negatively impacting model performance. Inspired by works like Long-CLIP \cite{zhang2024long}, our retrieval method employs stretching to extend the model's acceptable text length, enabling the retrieval of long text descriptions for environmental contexts.

\section{CVG-Text Dataset}

\begin{figure}[!t]
  \centering
  \includegraphics[width=0.45\textwidth]{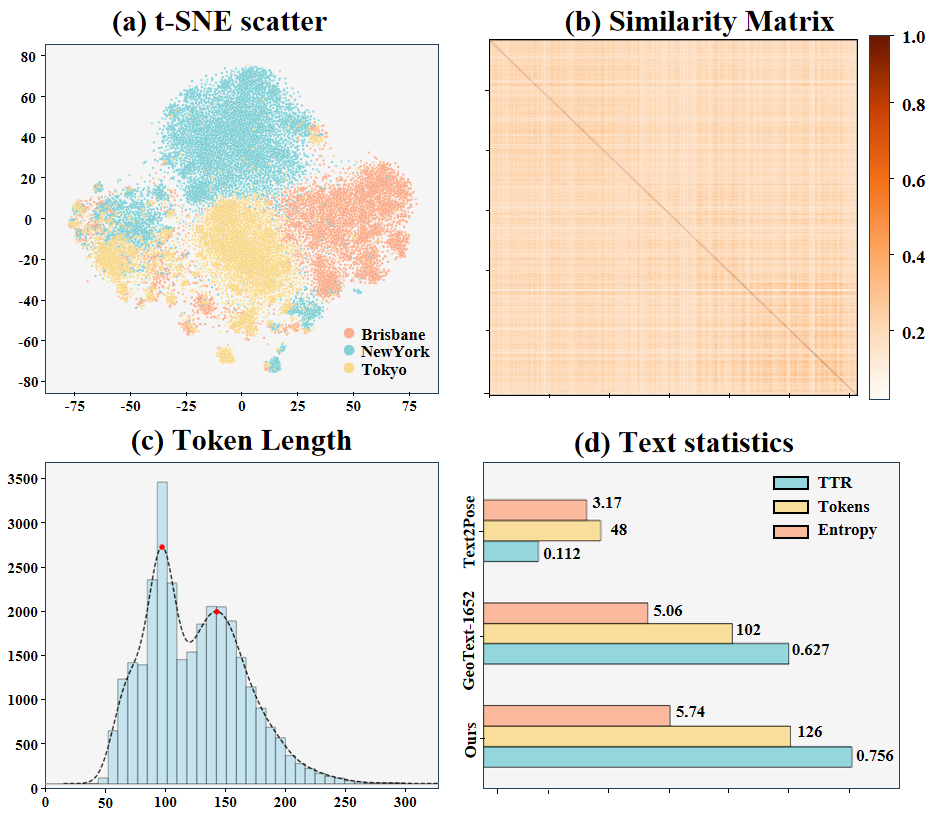}
  \caption{\textbf{Textual Feature Statistics Overview.} \textbf{(a)} t-SNE visualization of text data from different cities; \textbf{(b)} text similarity matrix; \textbf{(c)} token length distribution histogram; \textbf{(d)} comparison of text statistics across different datasets.
  }
  \vspace{-5mm}
  \label{fig:dataset}
\end{figure}

\subsection{Overview}

We introduce CVG-Text, a multimodal cross-view retrieval localization dataset designed to evaluate text-based scene localization tasks. CVG-Text covers three cities: New York, Brisbane, and Tokyo, encompassing over 30,000 scene data points. The data from New York and Tokyo is more oriented toward urban environments, while the Brisbane data leans towards suburban scenes. Each individual point includes corresponding street-view images, OSM data, satellite images, and associated scene text descriptions. The dataset is randomly split into training and test sets with a ratio of 5:1. More details can be found in the supplementary materials.

\noindent \textbf{Statistical Overview of Text Data.} Figure \ref{fig:dataset} presents the feature statistics of the scene text. The t-SNE dimensionality reduction visualization indicates a relatively dispersed distribution of the text data, reflecting high diversity. Texts from the same city exhibit clustering, while texts from different cities are clearly distinguishable, highlighting regional variations in text features. These differences are likely tied to the unique styles and cultural characteristics of each city. The text similarity matrix reveals low similarity, demonstrating the independence of the texts, which effectively represent each distinct scene and reduce the risk of confusion between different texts. The average text length generated by the multimodal large model GPT-4o exceeds 126 tokens, with two prominent peaks at 100 and 145 tokens, corresponding to single-view and panoramic images, respectively. Compared to datasets such as Text2Pose and GeoText-1652, our data shows superior performance in vocabulary richness, token length, and entropy, reflecting a higher quality of text.

\noindent \textbf{Comparisons with Existing Datasets.}
Table \ref{tab:datasets} compares CVG-Text with existing datasets. In contrast to common cross-view retrieval datasets, such as CVUSA \cite{workman2015wide} and VIGOR \cite{zhu2021vigor}, CVG-Text includes aligned text modality information, enabling the evaluation of text-based scene localization tasks and interpretability analysis for cross-view retrieval. Furthermore, CVG-Text demonstrates superior data completeness, encompassing panoramic street-views, single-perspective street-views, aerial images, and OSM data.

Compared to GeoText-1652 \cite{chu2023towards} dataset, which is primarily used for drone navigation, the text descriptions are directly derived from drone images and are used for drone image retrieval. Our task, however, focuses on addressing the needs of pedestrians, tourists, and other users, with text originating from street-view images and used for cross-domain retrieval of satellite or OSM images. There is a significant difference in both task objectives and the source of text. Furthermore, the coverage of drone images is more limited compared to the OSM and satellite images, making it difficult to achieve large-scale geo-localization.

\begin{table}[!t]
    \centering
    \setlength{\tabcolsep}{7.5pt}
    \footnotesize
    \caption{\textbf{Comparison of the proposed CVG-Text with existing cross-view datasets.} \# Ref. and \# G-Query represent the number of reference images and ground query images, respectively.}
    \begin{threeparttable}
     \begin{tabular}{p{2cm}@{\hskip 2mm}|@{\hskip 2mm}c@{\hskip 2mm}c@{\hskip 1.5mm}c@{\hskip 2mm}c@{\hskip 2mm}c@{\hskip 2mm}c}
        \toprule
         \textbf{Dataset} & \textbf{Text} & \textbf{Pano}\footnotemark[1] & \textbf{Single}\footnotemark[2] & \textbf{OSM} & \textbf{\# G-Query} & \textbf{\# Ref.}\\
        \midrule
        CVUSA \cite{workman2015wide} &  & \ding{51} &  &  & 44k & 44k \\
        CVACT \cite{liu2019lending} &  & \ding{51} &  &  & 128k & 128k \\
        VIGOR \cite{zhu2021vigor} &  & \ding{51} &  &  & 90k & 105k \\
        CVGlobal \cite{ye2024cross}&  & \ding{51} &  & \ding{51}  & 130K & 130K \\
        Uni.-1652 \cite{zheng2020university} &  &  & \ding{51} &  & 14K & 90k\\
        Geotext-1652 \cite{chu2023towards} & \ding{51} &  & \ding{51} &  & 14K & 90k \\
        \midrule
        Ours & \ding{51} & \ding{51} & \ding{51} & \ding{51} & 30k & 60k \\
        \bottomrule
    \end{tabular}
    \begin{tablenotes}
        \scriptsize 
        \item[1] Pano refers to the Panoramic street-view images.
        \item[2] Single refers to the Single-view street-view images.
    \end{tablenotes}
    \end{threeparttable}
    \label{tab:datasets}
    \vspace{-5mm}
\end{table}

\subsection{Data Collection}

We collected panoramic and single-perspective street-view images from different city areas using the Google Street View \footnote{\url{https://www.google.com/streetview/}} and Google Places API \footnote{\url{https://developers.google.com/maps/documentation/places/web-service}}. The resolution of the panoramic street-view images is \(2048 \times 1024\), with the north direction aligned to the middle column. The resolution of the single-perspective images is not fixed, but all are high-definition images. Based on the latitude and longitude coordinates of the street-view images, we collected corresponding satellite image tiles using Google Maps API\footnote{\url{https://www.google.com/maps}}. The size of the satellite images is \(512 \times 512\) , with a zoom level of 20 and a ground resolution of around 0.12m, aligned with the center of the street-view images.

Additionally, we collected corresponding OSM data for each region in vector format, encompassing global geographic and map information. OSM data contains numerous points of interest (POIs) with rich label information, such as restaurant names and bus stops, which are highly useful for geo-localization tasks. We utilize raster tiles provided by the OSM official website as retrieval targets, retaining various POI identifiers. The size of the OSM raster data is \(512 \times 512\), which is close to the image size and corresponding geographical extent of the satellite images. Since OSM data is dynamically updated, we collected street-view data from recent years to minimize discrepancies between the OSM and street-view data. Existing cross-view datasets typically include street-view data from before 2021; thus, rather than adding text annotations to existing datasets, we opted to collect new data.

\begin{figure}[t]
  \centering
  \includegraphics[width=0.5\textwidth]{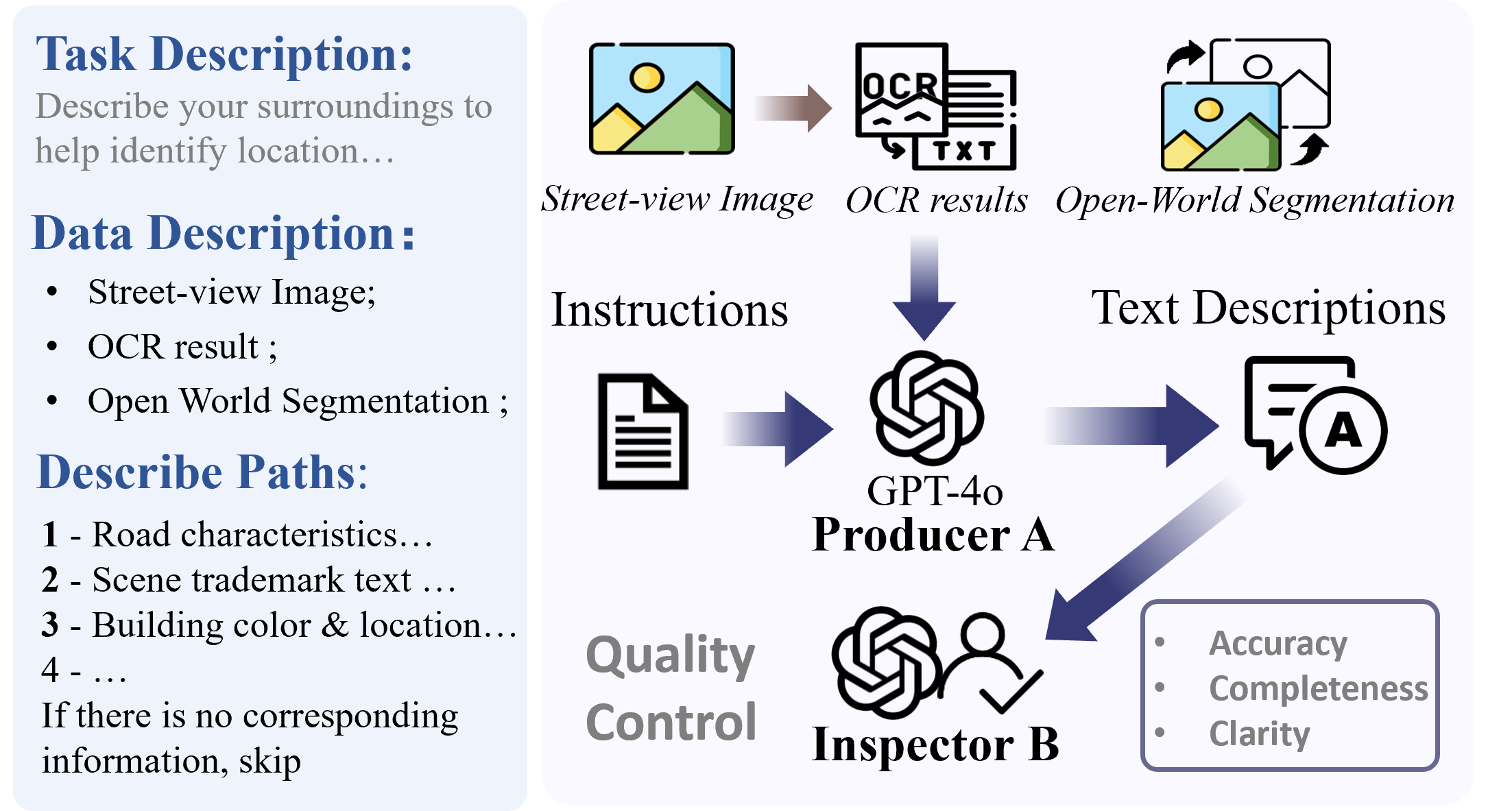}
  \caption{\textbf{Overall Process for Street-View Text Description Generation using GPT-4o.}
  }
  \vspace{-4mm}
  \label{fig:pipeline-dataset}
\end{figure}

\subsection{Text Data Synthesis}

Figure \ref{fig:pipeline-dataset} illustrates the main process of text synthesis based on GPT-4o. We first utilize Paddle OCR\footnote{\url{https://github.com/PaddlePaddle/PaddleOCR}} to capture text within street-view images, enabling GPT to focus on key semantic text information present in the images while reducing hallucination phenomena. Additionally, we perform open-world segmentation on the street-view images to provide more semantic details and location information for the subsequent text synthesis process. With the help of the semantic segmentation results, OCR outputs related to moving objects, such as vehicles, can also be filtered out. 

Next, we designed a systematic prompting scheme to generate scene text descriptions using GPT-4o. The system prompts are divided into four parts: task requirements, input data introduction, description paths, and response demands. Specifically, we employ a progressive scene analysis strategy to guide GPT in sequentially describing road features, building signage, and the overall environment, including aspects such as color, material, and distribution. In real-world applications, users may not know the exact geographic orientation. Therefore, text descriptions often use simple directional cues like front, back, left, and right to describe the scene. For more implementation details, refer to the supplementary materials.

In terms of quality control, we first filter out incorrectly formatted responses and conduct GPT-based review of both images and textual descriptions to ensure accuracy, consistency, content completeness, relevance, clarity, and comprehensiveness. Samples that do not meet the standards are re-synthesized. Additionally, we extracted 20\% of the samples (approximately 6,000) for manual expert review, involving 10 human evaluators and requiring around 100 hours of work. The pass rate for the manual checks was 77.6\%. Furthermore, street-view retrieval experiments using a retrained CLIP model showed that the Top-1 recall rate for text-to-street view was over 85.5\%, validating the high quality of the dataset's text descriptions.

\section{Method}

\subsection{Overview}
\label{overview}

\textbf{Problem Formulation.} In this work, we introduce a novel task for cross-view geo-localization with natural language. The objective of this task is to leverage natural language descriptions to retrieve its corresponding OSM or satellite images, of which the location information is usually available. 
The input for this task is a text description $Q$-$Text$ describing a street scene. 
In practical applications, users often have location prior, such as knowing they are in a particular district of New York City, but not the precise location. Therefore, the retrieval model utilizes this location prior to narrow the search scope $M$, querying a smaller subset of satellite images, $R$-$sat$, and OSM image, $R$-$OSM$.

\begin{figure}[t]
  \centering
  \includegraphics[width=0.5\textwidth]{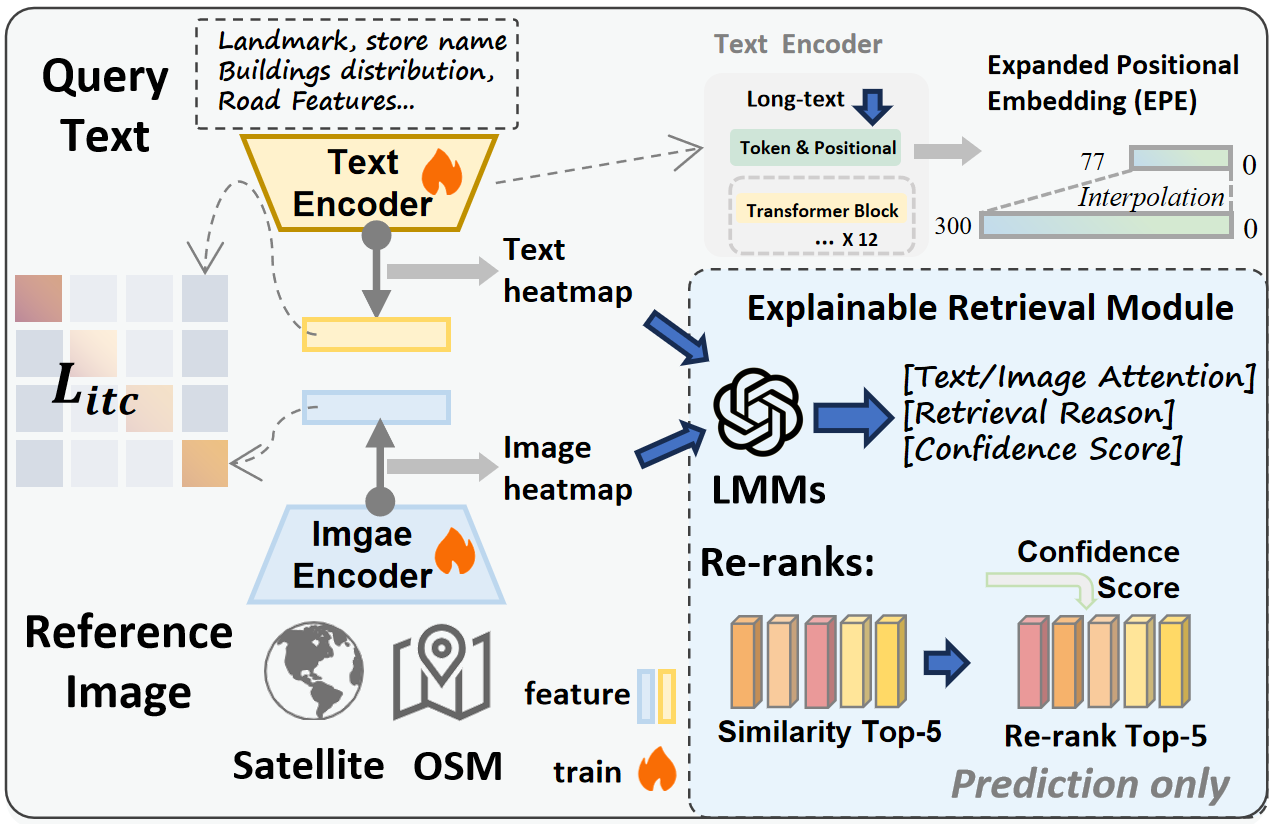}
  \caption{\textbf{The proposed CrossText2Loc method.} Street-view texts serve as query inputs, with satellite and OSM images as references. }
  
  \label{fig:model}
  \vspace{-4mm}
\end{figure}

\noindent \textbf{Model Architecture.} To address the challenge of the task, we propose the CrossText2Loc architecture (Figure \ref{fig:model}).It aligns the image and text domains through an enhanced text embedding module and a contrastive learning loss $L_{itc}$ enabling efficient text retrieval tasks. Furthermore, during inference, we introduce an optional Explainable Retrieval Module (ERM) to provide natural language analysis, enhancing the interpretability and confidence of retrieval decisions.

\begin{table*}[h]
    \centering
    \setlength{\tabcolsep}{4.5pt}
        \begin{tabular}{l|cccccc|cccccc}
            \toprule
            \multirow{3}{*}{\textbf{Method}} & \multicolumn{6}{c|}{\textbf{Satellite image}}  & \multicolumn{6}{c}{\textbf{OSM data}}\\
            \cmidrule(lr){2-7} \cmidrule(lr){8-13}
            & \multicolumn{2}{c}{\textbf{NewYork}} & \multicolumn{2}{c}{\textbf{Brisbane}} & \multicolumn{2}{c|}{\textbf{Tokyo}} & \multicolumn{2}{c}{\textbf{NewYork}} & \multicolumn{2}{c}{\textbf{Brisbane}} & \multicolumn{2}{c}{\textbf{Tokyo}} \\
            \cmidrule(lr){2-3} \cmidrule(lr){4-5} \cmidrule(lr){6-7} \cmidrule(lr){8-9} \cmidrule(lr){10-11} \cmidrule(lr){12-13}
            & \textbf{R@1} & \textbf{L@50} & \textbf{R@1} & \textbf{L@50} & \textbf{R@1} & \textbf{L@50} & \textbf{R@1} & \textbf{L@50} & \textbf{R@1} & \textbf{L@50} & \textbf{R@1} & \textbf{L@50} \\
            \midrule
            ViLT\cite{kim2021vilt} & 11.58 & 15.58 & 11.00 & 14.50 & 10.83 & 15.50 & 5.83 & 9.92 & 8.67 & 11.75 & 4.67 & 9.17 \\
            X-VLM\cite{zeng2022multi} & 15.74 & 16.86 & 15.67 & 17.60 & 12.46 & 14.34 & 16.14 & 17.26 & 20.46 & 21.94 & 9.53 & 10.94 \\
            SigLIP-B/16\cite{zhai2023sigmoid} & 19.67 & 21.08 & 19.58 & 22.00 & 15.58 & 17.92 & 20.17 & 21.58 & 25.58 & 27.42 & 11.92 & 13.67 \\
            SigLIP-SO400M\cite{zhai2023sigmoid} & 33.50 & 34.83 & 34.25 & 36.83 & 28.42 & 31.50 & 27.75 & 29.58 & 29.75 & 31.58 & 17.50 & 19.50 \\
            EVA2-CLIP-B/16\cite{fang2024eva} & 25.17 & 26.58 & 28.42 & 31.75 & 22.50 & 25.25 & 18.58 & 20.83 & 27.33 & 28.83 & 13.92 & 15.67 \\
            EVA2-CLIP-L/14\cite{fang2024eva} & 34.08 & 35.67 & 35.67 & 38.00 & 31.00 & 34.08 & 26.33 & 28.67 & 30.92 & 32.67 & 19.83 & 22.50 \\
            CLIP-B/16\cite{radford2021learning} & 26.67 & 28.17 & 29.92 & 32.58 & 24.00 & 27.25 & 27.42 & 29.42 & 30.83 & 32.67 & 17.75 & 19.92 \\
            CLIP-L/14\cite{radford2021learning} & 35.08 & 37.08 & 34.08 & 37.25 & 28.08 & 30.50 & 31.50 & 33.58 & 32.50 & 34.67 & 21.00 & 23.17 \\
            BLIP\cite{li2022blip} & 34.58 & 37.25 & 34.50 & 38.17 & 29.75 & 33.67 & 52.92 & 55.92 & 43.00 & 46.33 & 30.67 & 34.50 \\
            \midrule
            Ours (w/o ERM)  & \underline{46.25} & \underline{48.75} & \underline{43.58} & \underline{47.42} & \underline{36.83} & \underline{39.58} & \underline{59.08} & \underline{62.00} & \underline{46.08} & \underline{48.67} & \underline{34.33} & \underline{38.33} \\
            Ours  & \textbf{50.33} & \textbf{53.07} & \textbf{47.58} & \textbf{51.80} & \textbf{41.75} & \textbf{43.86} & \textbf{62.33} & \textbf{65.39} & \textbf{48.75} & \textbf{51.50} & \textbf{36.92} & \textbf{41.22} \\
            \bottomrule
        \end{tabular}
    \caption{\textbf{Quantitative comparison of different methods on CVG-Text.} R@1 represents the Top-1 image recall rate; L@50 represents the recall rate where localization error is less than 50 meters. [Key: \textbf{Best}, \underline{Second Best}]}
    \label{tab:main}
    \vspace{-4mm}
\end{table*}
\subsection{Image-text Contrastive Learning}

CrossText2Loc adopts a dual-stream architecture, consisting of a text encoder and a visual encoder that extract features from text queries and reference satellite or OSM images, respectively. Since the text describes scenes corresponding to street view images, there is a significant domain gap between these and the reference satellite and OSM data. As a result, the features obtained by the pre-trained encoder are spatially distant. Therefore, we align the image and text embedding spaces through contrastive learning, while jointly training both the text and image encoders. We set the batch size to $n$ and align the representations of images and text by minimizing the contrastive learning loss function $L_{itc}$. To achieve this, we uniformly encode \textit{panoramic} and \textit{single-view} texts as $t$, and \textit{satellite} and \textit{OSM} images as $v$. The loss function is expressed as:
\begin{equation}
L_{itc} = \sum_{i=1}^{n} \sum_{j=1}^{n} - \log \frac{ \exp \left( \text{sim}(v_i, t_j) / \tau \right) } { \sum_{k=1}^{n} \exp \left( \text{sim}(v_i, t_k) / \tau \right) }
\end{equation}
Where $v_i$ and $t_j$ represent the $i$-th image and the $j$-th text embedding vector respectively. $\tau$ is a learnable temperature used to control the sharpness of softmax distribution.

Scene description texts are generally complex and lengthy, and their matching with OSM or satellite images is not based on direct surface-level word matching, but requires deeper semantic understanding. However, previous text encoders, such as CLIP\cite{radford2021learning}, have limited positional embeddings, often resulting in truncation and insufficient global text representation. To address this issue, we adopt the \textbf{Expanded Positional Embedding (EPE)} method to extend the positional embeddings of the text encoder, which are then fed into the Transformer block. During subsequent embedding in the Transformer architecture, attention mechanisms are used to capture critical localization cues. Specially, we utilize linear interpolation to expand the positional embeddings to accommodate a sequence length of $N$ (300) tokens. The expanded positional embedding $P^*$ can be computed from the original positional embedding $P$ as follows:
\begin{small}
\begin{equation}
    P^*(x) = (1 - (x - \lfloor x \rfloor)) \cdot P(\lfloor x \rfloor) + (x - \lfloor x \rfloor) \cdot P(\lceil x \rceil)
\end{equation}
\end{small}
Where $P^*(x)$ represents the expanded positional embedding at position x, $P(\lfloor x \rfloor)$ and $P(\lceil x\rceil)$ are the values from the original positional embedding at the indices $\lfloor x \rfloor$ and $\lceil x \rceil$, with $\lfloor \cdot \rfloor$ and $\lceil \cdot \rceil$ denoting the floor and ceiling functions respectively. Due to the scene texts in our case are generated by GPT for simulating user data, there are no prominent short titles at the beginning which carry significant importance. Therefore, unlike LongCLIP \cite{zhang2024long} uses Knowledge Preserving Stretching, we use a full interpolation method.

\subsection{Explainable Retrieval Module (ERM)}
\label{subsec:ERM}

We also propose an optional Explainable Retrieval Module, designed to enhance the interpretability and trustworthiness of geolocation retrieval predictions through natural language. 

\noindent\textbf{Attention Heatmap Generation.}
Inspired by the method described in \cite{chefer2021generic}, we generate attention heatmaps to reveal the model's focus during retrieval. We initialize the correlation heatmap before the starting layer $s$ with an identity matrix $I$. Then, we iteratively process each attention layer from the starting layer $s$ to the output layer $L$. In each attention mechanism, we accumulate non-negative gradient contributions to generate attention heatmaps that highlight the regions the model focuses on. The image or text correlation heatmap $R$ can be calculated as:
\begin{small}
\begin{equation}
R^{(l)} = R^{(l-1)} + \frac{1}{H} \sum_{h=1}^{H} \max(0, \nabla A_h^{(l)} \odot A_h^{(l)}) R^{(l-1)}
\end{equation}
\end{small}
Where $A_{h}^{(l)}$ represents the attention weight of the $h$-th attention head at layer $l$, $\nabla A_{h}^{(l)}$ is the gradient of the attention weight, $\odot$ denotes the Hadamard product, $H$ is the total number of attention heads.

\noindent\textbf{LMM Interpretation.}
After generating the image and text heatmaps, we feed them into the LMM (GPT-4o) with a carefully designed prompt. First, the LMM observes the heatmaps to identify key clues in the image and text that the model focuses on during retrieval, such as specific landmarks or geographical features. Then, the LMM compares and reasons over the highlighted regions in the text and image, providing an explanation for why the model retrieved the query in a particular way, i.e., the rationale behind the matching of the query and reference data. Finally, the LMM outputs the confidence level of the retrieval rationale, simulating user decision-making.
LMMs are capable of providing retrieval explanations in natural language, beyond just similarity measures, which is highly beneficial for the interpretability and trustworthiness of geo-localization retrieval. 

\noindent\textbf{Confidence re-ranking.} The confidence scores obtained through explainable retrieval can be used to re-rank the original top-5 similarity scores from the retrieval match. We set the results with top-1 confidence scores lower than 0.5 and normalize both the similarity and confidence scores to the [0, 1] range, summing them to obtain the re-ranked Top-5 results. In the subsequent experimental section, we demonstrate that the effective re-ranking strategy helps simulate user retrieval decisions and improves recall in tasks.

\section{Experiment}
\subsection{Experimental Setup}

\textbf{Implementation Details.} We used the CLIP-L/14@336px model \cite{radford2021learning} pre-trained by OpenAI as the backbone, with the Adam optimizer \cite{kingma2014adam}, a learning rate of 1e-5, and cosine learning rate decay. The batch size was set to 128, and training was conducted over 40 epochs on four NVIDIA A100 GPUs. The image resolution was set to default \(336 \times 336\), and the text context length $N$ was 300 tokens. We initialized the temperature coefficient $\tau$ from the checkpoint. 
Moreover, as mentioned in Section \ref{overview}, based on practical application requirements and utilizing user location priors, we set the retrieval range $M$ to 100. During testing, the reference database consists of 100 samples from the nearby area, covering an area of approximately 10 km². 
$M$ is a configurable parameter, and we present additional results with different $M$ settings in the supplementary material. The image resolution for all comparison methods follows their respective default best settings, and all methods are trained on the same proposed dataset.

\begin{figure*}[tb]
  \centering
  \includegraphics[width=\textwidth]{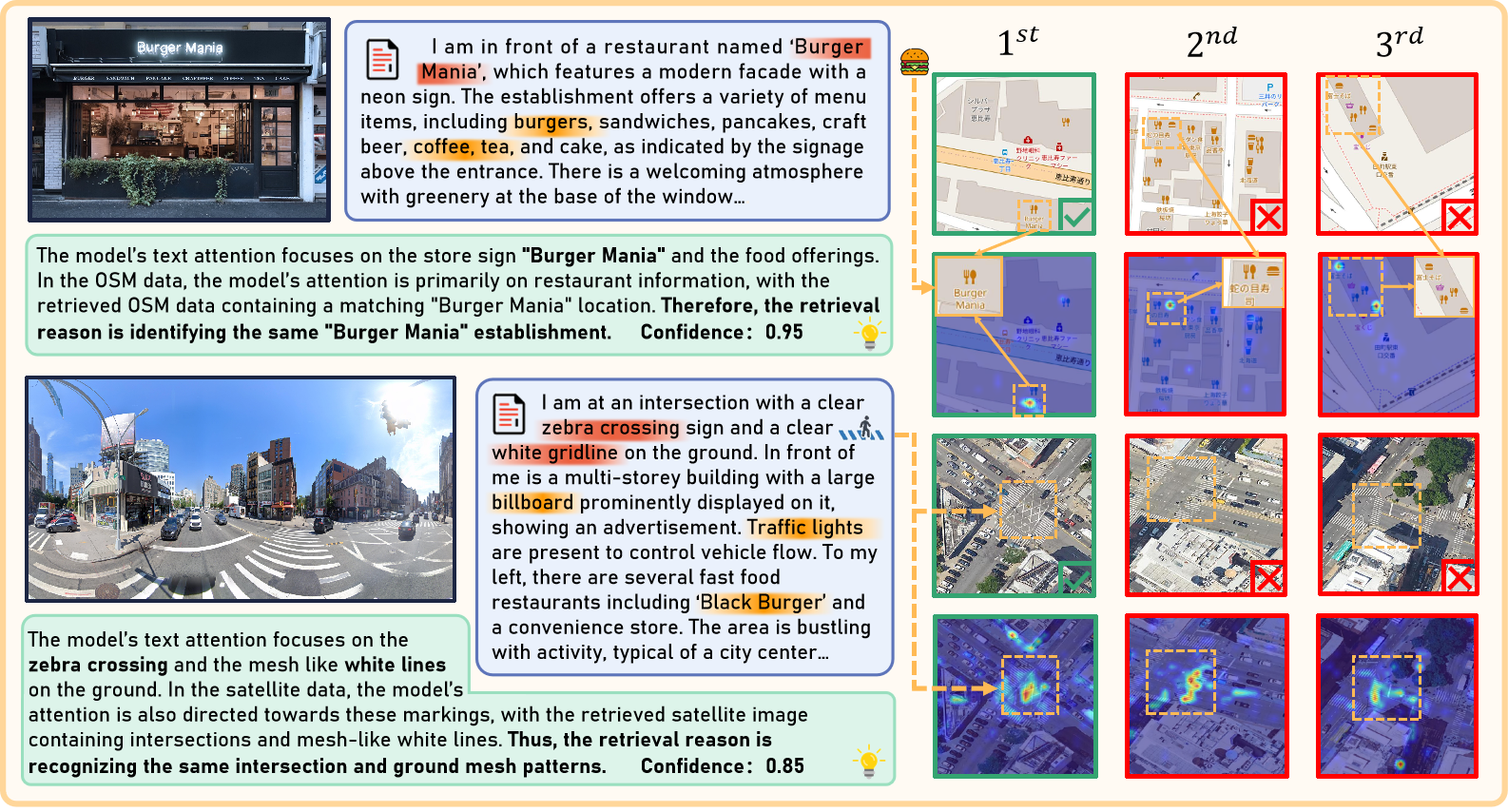} 
  \caption{\textbf{Qualitative retrieval results
on CVG-Text Dataset.}  The left side of the figure displays the original street-view data, synthetic text data with corresponding response heatmaps, and retrieval reason provided by our ERM module. The right side shows the top three retrieval results with corresponding response heatmaps; green indicates correct matches and red denotes incorrect results.}
  \vspace{-1em}
  \label{fig:Qualitative}
\end{figure*}

\noindent \textbf{Evaluation Metrics.} Following previous cross-view geo-localization works \cite{shi2019spatial,Deuser_2023_ICCV}, we use the image recall accuracy of the top K images as an evaluation metric to assess text retrieval localization performance. Specifically, given a query text for a certain location, if its ground-truth OSM and satellite image is within the top K retrieval results, the location is considered “successfully localized.” Additionally, we also provide the localization recall rate metric \cite{xia2024text2loc,kolmet2022text2pos}. Similarly, the localization recall rate refers to the proportion of retrieved results where the distance to the actual location is below a specified threshold.

\subsection{Geo-localization Performance}

We evaluated the performance of various text-based retrieval methods under different settings of satellite images and OSM data, with the results shown in Table \ref{tab:main}. Among the existing approaches, BILP achieves optimal performance, as it is not constrained by limitations on text embedding length.Our method, even without the ERM, achieved the best results. Compared to the baseline CLIP method, the proposed approach improved Recall@1 by 14.1\% and Recall@10 by 14.8\%, demonstrating its advantages in this task.

Next, we evaluated the performance of methods in different cities and across different data sources (Satellite/OSM). In New York, OSM outperformed satellite imagery due to richer POI data, such as bus stops and store names, which are hard to identify in satellite images. In contrast, the level of detail in OSM data for Brisbane is limited, and in this case, the localization performance of satellite images is comparable to that of OSM. In Tokyo, due to the poor pre-training of CLIP on Japanese, the model's response to certain Japanese words in the street view description text and Japanese POIs in OSM data is weak, leading to the least favorable performance in Tokyo. The supplementary materials also include cross-city evaluations and a collaborative retrieval method for OSM and satellite images.

\subsection{Explainable Retrieval}

We present the results of the Explainable Retrieval Module (ERM) in Figure \ref{fig:Qualitative}. This method uses text and the heatmap responses of the retrieved images to highlight the key features the localization model focuses on. In the first example, the model focuses on “Burger Mania”. Interestingly, even in non-top-1 results, it still emphasizes the burger icon. In the second example, the model focuses on the “zebra crossing” and “white gridline”, with the first three retrieval results all highlighting the zebra crossing, and the best retrieval result matching both features. We provide a quantitative evaluation of the ERM module’s retrieval rationale in the supplementary materials, using similarity matching with human-written explanations, CIDEr\cite{Vedantam_2015_CVPR}, ROUGE\_{L}\cite{lin2004rouge}, and multidimensional human scoring.

Moreover, the confidence scores obtained from LMMs essentially simulate the user's confidence decision-making process. When the confidence score is low, it indicates that, even though the similarity score of the result may be high, it is still not convincing. In this case, by applying the re-ranking strategy, we can obtain better retrieval results, as shown in Table \ref{tab:main}. The practical value of the Explainable Retrieval Module (ERM) lies in assisting users to assess the rationale behind the localization decision. They can select the most reasonable result from similar candidates, whereas previous methods only relied on the highest similarity match. If the rationale provided by ERM is not convincing enough, users can choose to re-rank the remaining search results or add additional descriptions to provide more visual clues.

\begin{figure}[t]
  \centering
  \includegraphics[width=0.5\textwidth]{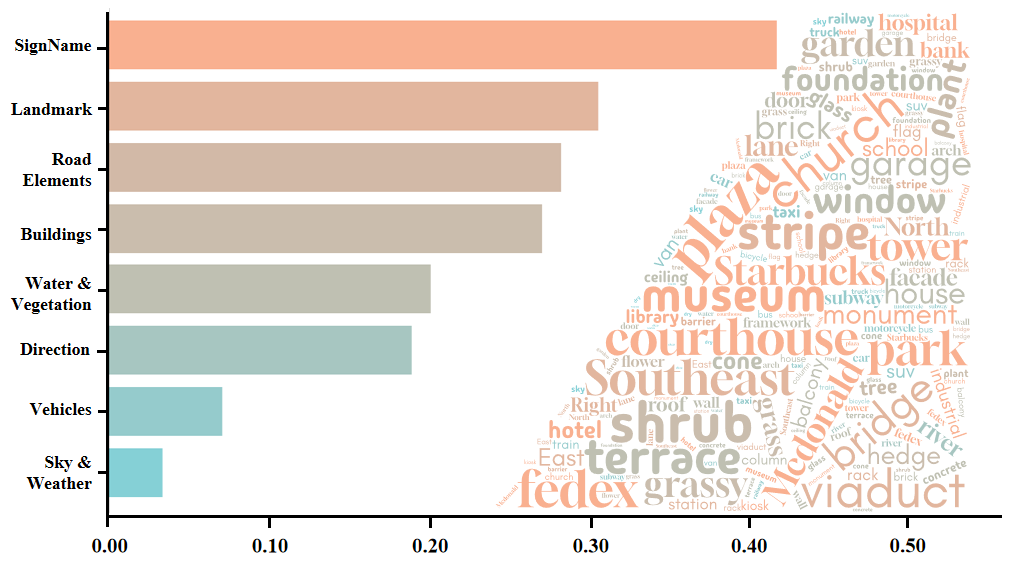}
  \vspace{-1em}
  \caption{The average attention score for each object category (\textit{left}) and the word cloud of selected attention words (\textit{right})\textbf{.}
  }
  \vspace{-1em}
  \label{fig:Score}
\end{figure}

Additionally, we generate feature heatmap responses based on text and reference images. By analyzing the top-ranked words in the model’s attention, we identified eight focal subcategories, such as "LandMarker" for landmarks and "SignName" for store names or bus stop names. A list of these words and categories is in the supplementary materials. Attention scores for each category in the scene are shown in Figure \ref{fig:Score}. Results reveal that the cross-view retrieval model mainly focuses on landmarks and road information, with less attention to vehicles, sky, and weather, as they are less relevant for localization. This highlights key clues for understanding cross-view retrieval models.

\subsection{Ablation Studies}

We evaluated the text synthetic effects of those directly generated by GPT versus  those generated through the integrated process of CVG-Text, as shown in Table \ref{tab:Auc}. CVG-Text exhibits higher text length and vocabulary complexity (TTR), with lower text similarity, indicating better text quality. The inclusion of OCR assistance helps precisely capture textual details in street view images, reducing GPT’s tendency for vague descriptions and hallucinations, and effectively generates text with key localization cues, significantly improving OSM retrieval performance. Open-World Segmentation further enhances GPT’s semantic and spatial understanding, improving satellite image retrieval performance.

\begin{table}[!t]
\centering
\setlength{\tabcolsep}{1.5mm}
\begin{tabular}{@{}c c c c c c@{}}
\toprule
\multirow{2}{*}{\textbf{Methods}} & \multicolumn{3}{c}{\textbf{Text quality}} & \multicolumn{2}{c}{\textbf{Retrieval result}} \\
\cmidrule(lr){2-4} \cmidrule(lr){5-6}
 & \textbf{Len} & \textbf{TTR} & \textbf{Simi.} & \textbf{R@1-OSM} & \textbf{R@1-Sat} \\
\midrule
Baseline & 108 & 0.74 & 0.22 & 25.17 & 38.00 \\
Ours & 126 & 0.76 & 0.17 & 59.08 & 46.25 \\
\midrule
$\Delta$ & +18 & +0.02 & -0.05 & +33.91 & +8.25 \\
\bottomrule
\end{tabular}
\caption{Comparison between Baseline (directly using GPT-generated Text) and our CVG-Text dataset in New York.}
\label{tab:Auc}
\vspace{-4mm}
\end{table}


\begin{table}[!t]
\centering
\setlength{\tabcolsep}{9.5pt}
\begin{tabular}{@{}lcccc@{}}
\toprule
\multirow{2}{*}{\textbf{Method}} & \multicolumn{2}{c}{\textbf{Satellite}} & \multicolumn{2}{c}{\textbf{OSM}} \\
\cmidrule(lr){2-3} \cmidrule(lr){4-5}
& \textbf{R@1} & \textbf{R@5} & \textbf{R@1} & \textbf{R@5} \\
\midrule
SigLIP & 19.67 & 53.33 & 20.17 & 41.83 \\
SigLIP + EPE & \textbf{29.50} & \textbf{64.00} & \textbf{45.25} & \textbf{70.25} \\
\midrule
CLIP & 35.08 & 71.42 & 31.50 & 55.42 \\
CLIP + EPE & \textbf{46.25} & \textbf{81.58} & \textbf{59.08} & \textbf{82.75} \\
\bottomrule
\end{tabular}
\caption{Ablation Study of Expanded Positional Embedding (EPE) Module in New York.}
\label{tab:ablation_study_EPE}
\vspace{-4mm}
\end{table}

\begin{table}[h]
    \scriptsize
    \centering
    \renewcommand{\arraystretch}{1}
    \setlength{\tabcolsep}{3pt}
    \resizebox{\columnwidth}{!}{
        \begin{tabular}{l|cc|cc|c}
            \toprule
            \multirow{2}{*}{Method} & \multicolumn{2}{c|}{OSM} & \multicolumn{2}{c}{Satellite} & R@1 \\
             & R@1 & R@10 & R@1 & R@10 & average \\
            \midrule
            SAFA \cite{shi2019spatial} & 19.25 & 44.88 & 77.40 & 95.30 & 48.32 \\
            Geo-Dtr \cite{zhang2023cross} & 24.10 & 53.30 & 86.45 & 98.80 & 55.28 \\
            Sample4G \cite{Deuser_2023_ICCV}  & 27.10 & 62.90 & \underline{91.70} & \underline{99.20} & 59.40 \\
            Text-only & \underline{59.08} & \underline{90.00} & 46.25 & 91.00 & 52.67 \\
            Sample4G \cite{Deuser_2023_ICCV} +Text & \textbf{67.30} & \textbf{97.20} & \textbf{98.40} & \textbf{99.80} & \textbf{82.80} \\
            \bottomrule
        \end{tabular}
    }
    \caption{Impact of Adding Text Branch for Cross-View Retrieval.}
    \label{tab:sample4g}
    \vspace{-4mm}
\end{table}

We employed two different learning architectures, CLIP \cite{radford2021learning} and SigLIP \cite{zhai2023sigmoid}, to validate the contribution of our proposed Expand Positional Embedding (EPE) module. As shown in Table \ref{tab:ablation_study_EPE}, the use of the EPE module significantly improves the performance of both text retrieval methods on satellite images and OSM, with R@1 recall rates increasing by 10.5\% and 26.3\%, respectively. This demonstrates the effectiveness of extending text encoding length for handling longer text content in this task.

\subsection{Further Discussion on Cross-view retrieval task}

Previous cross-view retrieval tasks have primarily focused on image-to-image queries. Leveraging the CVG-Text dataset, we expanded this task to multimodal image-text joint queries. Building on the state-of-the-art cross-view street-view retrieval method, Sample4G \cite{Deuser_2023_ICCV}, we added an additional text query branch via fusing the similarity scores. We conducted satellite and OSM retrieval experiments in the New York area of the CVG-Text dataset. As shown in Table \ref{tab:sample4g}, compared to using only street-view images as queries, the addition of a text branch improved Top-1 recall by 6.7\% for the satellite retrieval task and by 40.2\% for the OSM retrieval task, effectively enhancing retrieval accuracy. Since the detailed scene descriptions in CVG-Text align more closely with POI data in OSM, the improvement for OSM retrieval is more pronounced than for satellite image data.

\section{Conclusion}

In this work, we explore the task of cross-view geo-localization using natural language descriptions and introduce the CVG-Text dataset, which includes well-aligned street-views, satellite images, OSM images, and text descriptions. We also propose the CrossText2Loc text retrieval localization method, which excels in handling long-text retrieval and interpretability for this task. This work represents another advancement in the field of natural language-based localization. It also introduces new application scenarios for cross-view localization, encouraging subsequent researchers to explore and innovate further.

{
    \small
    \bibliographystyle{ieeenat_fullname}
    \bibliography{main}
}

\end{document}